%% file: root.tex
\documentclass[letterpaper, 10 pt, conference]{ieeeconf}  %

\IEEEoverridecommandlockouts                              %

\usepackage[pagebackref,breaklinks,colorlinks,citecolor=cvprblue]{hyperref}
\usepackage{amssymb}
\usepackage{array}
\usepackage{mathtools}
\usepackage{leftindex}
\usepackage{tabularray}
\usepackage{xcolor}
\usepackage{gensymb}
\usepackage{indentfirst}
\usepackage{caption}
\usepackage[most]{tcolorbox}

\usepackage[nocompress]{cite}

\newcommand{\insertfig}{\vspace{15pt}\includegraphics[width=\linewidth]{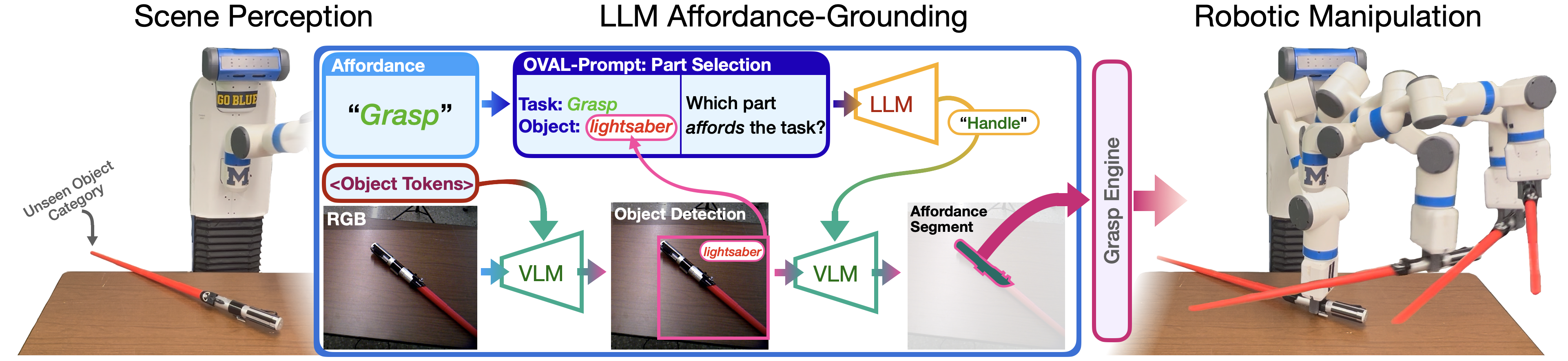}\captionof{figure}{Overview of Affordance-Prompting, a pipeline for open-vocabulary affordance segmentation and localization. Left: Given an RGB image and list of object tokens, a pre-trained Vision Language Model (VLM) provides object detections. Middle: The detected object is composed with a natural language affordance to form an OVAL-Prompt that is fed into a pre-trained Large Language Model (LLM) to identify which part of the detected object supports the task affordance. Right: The identified object part is fed back into the VLM to generate a part-affordance-segment that is used by a grasp engine (e.g. Dex-Net~\cite{mahler2017dexnet}) to enable robot manipulation.}\vspace{-12pt}}%

\makeatletter
\apptocmd{\@maketitle}{\centering\insertfig}{}{}%
\makeatother

\long\def\sl#1{\textcolor{orange}{\bf \small [SL: #1]}}

\title{\LARGE \bf
OVAL-Prompt: Open-Vocabulary Affordance Localization\\for Robot Manipulation through LLM Affordance-Grounding
}

\author{Edmond Tong$^{1}$, Anthony Opipari$^{1}$, Stanley Lewis$^{1}$, Zhen Zeng$^{2}$, and Odest Chadwicke Jenkins$^{1}$%
\thanks{$^{1}$E. Tong, A. Opipari, S. Lewis, and O.C. Jenkins are with the Department of Robotics, University of Michigan, Ann Arbor, MI, USA, 48109.}
\thanks{$^{2}$Z. Zeng is with J.P. Morgan AI Research.}
\thanks{$^{*}$E. Tong is the corresponding author:
        {\tt\small ekjt@umich.edu}}%
}

\begin{document}

\maketitle

\input{secs/0_abstract}
\input{secs/1_introduction}

\input{secs/2_related}

\input{secs/3_problem}
\input{secs/4_approach}

\input{secs/5_experiments}

\input{secs/6_conclusion}

\section*{Acknowledgment}

This work was supported in part by a Qualcomm Innovation Fellowship, J.P. Morgan AI Research, Amazon, and Ford Motor Company.

\bibliographystyle{IEEEtran}  
\bibliography{bib}

\clearpage

\input{secs/7_supplementary}

\end{document}

%% file: secs/0_abstract.tex
\begin{abstract}
In order for robots to interact with objects effectively, they must understand the form and function of each object they encounter.
Essentially, robots need to understand \textit{which} actions each object affords, and \textit{where} those affordances can be acted on.
Robots are ultimately expected to operate in unstructured human environments, where the set of objects and affordances is not known to the robot before deployment (i.e. the open-vocabulary setting).
In this work, we introduce OVAL-Prompt, a prompt-based approach for open-vocabulary affordance localization in RGB-D images.
By leveraging a Vision Language Model (VLM) for open-vocabulary object part segmentation and a Large Language Model (LLM) to ground each part-segment-affordance, OVAL-Prompt demonstrates generalizability to novel object instances, categories, and affordances without domain-specific finetuning. 
Quantitative experiments demonstrate that without any finetuning, OVAL-Prompt achieves localization accuracy that is competitive with supervised baseline models.
Moreover, qualitative experiments show that OVAL-Prompt enables affordance-based robot manipulation of open-vocabulary object instances and categories.
 
\end{abstract}

%% file: secs/1_introduction.tex
\section{Introduction}

For robots to function effectively in unstructured settings like homes and offices, they must be adept at identifying objects in their surroundings and utilizing them appropriately.
The potential uses for each object can be understood in terms of the object's ``affordances."
Originally termed by psychologist James J. Gibson~\cite{gibson_ecological_2015}, an object's affordance describes what actions are possible for a robot to perform using the given object -- thereby enabling the robot to reason about effective object use. 
For example, knives afford cutting solids and spoons afford scooping liquids.
    Once the robot has detected which object affordances are necessary for task completion, it must then be able to localize the relevant object geometry that enables said affordances.
For this work, we refer to the combined detection and localization task as `affordance grounding.'
Traditional approaches to affordance grounding have relied heavily on extensive labeled datasets and task-specific training, limiting their flexibility and scalability for robotic applications.
Roboticists have increasingly sought to address this limitation by developing approaches for open-vocabulary settings, where the set of objects~\cite{Learningluo} and affordances~\cite{3d_open_aff} is not constrained to the same fixed set that was used during training.
In particular, Large Language Models (LLMs) and Vision Language Models (VLMs) have been proposed as sources of knowledge to address the open-vocabulary setting~\cite{saycan2022arxiv,3d_open_aff,qian2024affordancellm}.

There is a growing interest in using LLMs and VLMs for robotic applications ~\cite{codeaspolicies2022,saycan2022arxiv,kawaharazuka2024real,kambhampati2024llms}, with notable examples including CLIP~\cite{clip,lerftogo2023} and Say-Can~\cite{saycan2022arxiv}.
Specifically, LLMs and VLMs have been proposed as knowledge bases that can be tapped into for solving classical robotics challenges.
For example, within planning, Valmeekam et al. and Kambhampati et al. proposed using LLMs as a source of knowledge within `LLM-Modulo' planning frameworks but find that LLMs struggle to plan in isolation~\cite{valmeekam2024planning,kambhampati2024llms}.
For affordance grounding and localization specifically, Qian et al. propose finetuning a LLM to generate affordance mask-tokens, which can be decoded into localized affordance segments~\cite{qian2024affordancellm}.
Nguyen et al. propose finetuning a text-encoder to be used for open-vocabulary feature correlations within an affordance localization pipeline~\cite{3d_open_aff}.
In contrast, we set out to understand the potential for pre-trained LLMs and VLMs to be used without domain-specific finetuning.

Our work is motivated by two simple questions: \textit{Can pre-trained LLMs and VLMs be used to ground object-part affordances without domain-specific finetuning and do they enable robots to manipulate open-vocabulary affordances?}
To answer these questions, we develop an LLM and VLM-based affordance-grounding pipeline to detect and localize part-based object affordances in images.
By using pre-trained foundation models, the developed pipeline is applicable to the open-vocabulary setting where object instances, categories, and affordances are evaluated without having been trained for in a supervised fashion.
The proposed pipeline is quantitatively evaluated through multiple experiments to measure its accuracy relative to state-of-the-art supervised methods.
Furthermore, real robot experiments are performed using the affordance-grounding pipeline to establish whether the pipeline is applicable to real-world robotic scenarios.

In summary, this paper sets out to advance affordance-based robotic manipulation with the following contributions:

\begin{enumerate}
    \item First, the OVAL-Prompt algorithm is introduced to perform open-vocabulary affordance localization. OVAL-Prompt uses a LLM to ground part-segments generated by a VLM with corresponding action-affordances.
    \item Second, we explore the importance of prompt structure in affordance-prompting to ground visual-affordances with natural language tokens. Results from these experiments suggest using a LLM to translate affordance to object part and prompt it for similar names.
    \item Third, we demonstrate that OVAL-Prompt, which uses no domain finetuning, is competitive with multiple finetuned state-of-the-art baseline models on a real-world affordance localization dataset.
    \item Fourth, robot experiments are performed showing OVAL-Prompt can be successfully used on real robot platforms for successful affordance-based object and tool grasping.%
\end{enumerate}

%% file: secs/2_related.tex
\section{Related Works}

\subsection{Closed-Vocabulary Affordance Localization}
Early approaches to affordance grounding relied on hand-crafted features such as geometric shapes \cite{geometric_umd}. However, recent research has shifted towards deep learning techniques \cite{aff_dl_survey}. Traditional deep learning approaches use a convolutional backbone (e.g. VGG16~\cite{simonyan2014very}) for feature extraction and use the resulting latent features to generate a final prediction~\cite{CNN-rgb,nguyen2017object,do2018affordancenet,nagarajan2019grounded,fang2020learning,deng20213d}.
These approaches rely on supervised training with a fixed set of object affordances that do not change during inference, thereby limiting their applicability to robotic applications.
In contrast, the present paper sets out to address affordance localization in the open-vocabulary setting where the object categories and affordances are not fixed after training.

\subsection{Open-Vocabulary Affordance Localization}

`Open-vocabulary' refers to machine learning tasks in which models are trained using a fixed set of category labels and evaluated using a non-fixed set of category labels~\cite{wu2023open}.
In the context of affordance localization, the open-vocabulary setting has received increasing interest~\cite{Learningluo,li2023locate,3d_open_aff,li2023oneshot}.
Luo et al. introduce a large-scale image dataset for open-vocabulary affordance localization and develop a feature co-relation strategy to achieve open-vocabulary inference \cite{Learningluo}.
Li et al. propose a vision transformer and exocentric human demonstrations to identify affordances in egocentric images with weakly-supervised training~\cite{li2023locate}.
Similarly, Li et al. propose a vision transformer trained with a one-shot approach in which only one labeled example per object category is used for training~\cite{li2023oneshot}. 
Nguyen et al. focus on affordance localization within 3D point clouds with a contributed dataset and use feature correlations along with a pre-trained text-encoder for the open-vocabulary setting~\cite{3d_open_aff}.
In contrast, the present paper sets out to avoid expensive finetuning strategies and instead to explore the potential for pre-trained foundation models to be used for open-vocabulary affordance localization.

\subsection{Foundation Models}

Several Large Language Models (LLMs), such as GPT-4\cite{OpenAIGPT4} and Gemini\cite{GeminiDeepMind}, have demonstrated exceptional capabilities in tasks like test-taking, gaining significant traction in both research and practical applications. These advancements in LLMs, have paved the way for their widespread adoption across various domains. In the field of robotics, LLMs are increasingly being utilized for a broad spectrum of functions. These include enhancing human-robot interaction \cite{saycan2022arxiv}, planning\cite{song2023llmplanner}, and navigation\cite{shah2022lmnav}. 

This work aims to answer the questions: Do LLMs understand affordances? While LLMs are known for their remarkable recall and comprehension abilities, they have also been criticized for concocting infeasible plans\cite{llmplan}, and demonstrating flawed reasoning\cite{Kambhampati2023LLMs}.

%% file: secs/3_problem.tex
\section{Affordance Localization}

\subsection{Problem Definition}

We tackle zero-shot open-vocabulary affordance localization, by identifying actionable object parts from images and descriptions without prior specific training. We focus on the task of open-vocabulary affordance localization in images. Given a RGB image, $I\in \mathcal{R}^{H\times W\times 3}$, the goal is to detect and segment the affordance labels, $\mathcal{L}=\{l_1, l_2, \ldots, l_N\}$, present in the image. 
Crucially, for the open-vocabulary setting, the set of test labels can differ from the one used during model training.

\subsection{Evaluation Metric}
To evaluate zero-shot affordance localization, we use the weighted F1-score defined by \ref{eqn:Fscore} that has been used in previous affordance localization papers \cite{geometric_umd}.
\begin{equation}
\label{eqn:Fscore}
        F_{\beta}^{w} = \frac{(1 + \beta^2) \cdot P_{r}^{w} \cdot R_{c}^{w}}{\beta^2 \cdot P_{r}^{w} + R_{c}^{w}}, \text{ with } \beta = 1
\end{equation}
Where  $P_{r}^{w}$  and  $R_{c}^{w}$  are weighted versions of the standard precision $P_{r} = \frac{TP}{TP + FP} $  and recall $R_{c} = \frac{TP}{TP + FN} $measures. Where TP = true positive, FP = false positive, FN = false negative. Precision indicates how many of the items identified as relevant are actually relevant, while recall measures how many relevant items were identified out of all relevant items. For more information see \cite{geometric_umd}. 
In zero-shot evaluation, an affordance that is not predicted but exist in the ground truth sample is assigned an F-score of 0.%

%% file: secs/4_approach.tex
\section{OVAL-Prompt: Affordance-Prompting}

Our method processes a RGB image and task description, using a VLM to detect and list objects within the image. This list informs a natural language prompt for a LLM, which specifies the object part related to the task. The VLM then segments this part, creating a mask for manipulation tasks. If segmentation fails, the LLM is reprompted for alternative part names, and the VLM tries again.

This approach is highly adaptable, using off-the-shelf components. The LLM and VLM are interchangeable, allowing for easy updates in the future, promising significant enhancements as LLMs and VLMs progress.

\subsection{VLM}
We employ the Vision-Language Model (VLM) for object detection and part segmentation. Initially, we supply the VLM with an image and a list of potential objects to identify, discarding detections below 50\% confidence. This filtered list is then provided to the LLM where it is processed and returns an object part. Upon receiving an object part from the LLM, the VLM segments it and generates a binary mask.

We chose VLpart\cite{vlpart} for its part segmentation capability, specifically the ``swinbase\_cascade\_lvis\_paco \_pascalpart\_partimagenet\_in" model, because of its high mAP score in their evaluation.

\subsection{LLM}
Current research highlights the significance of crafting precise prompts for Large Language Models (LLMs). Our method uses straightforward natural language prompts with organized outputs, engaging LLMs at three key stages: identifying relevant objects, pinpointing the object part related to the task, and seeking alternative part names. 

First, we provide the LLM with a list of items and the task, from which it selects objects suited to the task.

\begin{tcolorbox}[colback=blue!5!white,colframe=blue!75!black,title=Object Prompt]
\noindent Your task is to $<$task$>$. Which of these objects can do this task: $<$list of objects$>$ \\
Answer like the following:

Objects:\\
Reason:
\end{tcolorbox}

Our prompts for the LLM include ``Object" and ``Reason" sections to specify the chosen object and its justification, enhancing response validity through a logical ``chain of thought."
Following object identification, the LLM is further queried for the task-related object part, using a similar prompt structure. If the part segmentation by the VLM fails, we seek alternative part names from the LLM, like ``cup side" for ``cup body." We use GPT-4\cite{OpenAIGPT4} API with a zero temperature setting for consistent outputs.

%% file: secs/5_experiments.tex
\section{Experiments}

\subsection{Affordance Localization}

In our experiments, we aim to determine the LLM's understanding of affordances by testing our pipeline on the UMD\cite{geometric_umd} dataset.

We selected the UMD\cite{geometric_umd} dataset for its extensive usage and the availability of numerous benchmarks for comparison. This dataset features RGB-D images with ground truth pixel-wise affordance labels for 105 items encompassing different viewpoints, 7 affordances, and 17 object classes. We use the weighted F-score as used in the original publication.

The UMD\cite{geometric_umd} dataset, includes the following objects [knife, saw, scissors, shears, scoop, spoon, trowel, bowl, cup, ladle, mug, pot, shovel, turner, hammer, mallet, tenderizer] and affordances [grasp, cut, scoop, contain, pound, support, wrap-grasp]. We did not finetune or train on this list, but we use the list of objects and affordances as inputs to the system. If a given object was not detected or an object part was not detected, a f-score of 0 was assigned to that case. We ran this method on the test set images and the results are in table \ref{table:umd_performance}.

\begin{table*}[ht!]
\centering
\caption{Performance on UMD Dataset}
\label{table:umd_performance}
\begin{tabular}{lcccccccc}
\hline
 & HMP\cite{geometric_umd} & DeepLab\cite{chen2017deeplab} & AffordenceNet\cite{do2018affordancenet} & Attention-CNN\cite{gu_visual_2021} & RelaNet\cite{zhao_object_2020} & GSE\cite{GSE} & OVAL-Prompt(Ours)\\ 
 
 \hline
Grasp & 0.367 & 0.620 & 0.731 & \textbf{0.784} & 0.737 & 0.779 & 0.650 \\
W-Grasp & 0.373 & 0.730 & 0.814 & 0.822 & 0.824 & \textbf{0.840} & 0.718 \\
Cut & 0.415 &0.600 & 0.762 & 0.761 & 0.755 & 0.776 & \textbf{0.823}\\
Contain & 0.810 & 0.900 & 0.833 & 0.840 & \textbf{0.941} & 0.924 & 0.688 \\
Support & 0.643 & 0.600 & 0.821 & 0.844 & 0.866 & \textbf{0.893} & 0.538 \\
Scoop & 0.524 &0.800 & 0.793 & \textbf{0.862} & 0.802 & 0.856 & 0.753 \\
Pound & 0.767 & 0.880 & 0.836 & 0.847 & 0.879 & \textbf{0.918} & 0.809\\
\hline
Average &0.557 & 0.733 & 0.799 & 0.823 & 0.829 & \textbf{0.855} & 0.711 \\
\hline
\end{tabular}
\end{table*}
OVAL-Prompt achieves a F-score +0.154 higher than HMP on average but -0.144 lower than the highest-performing GSE model.  Despite its sub state-of-the-art performance it achieves a level of performance that is functional for real-world application.

\subsubsection{Failure mode}

The segmentation from the VLM resulted in lower F-scores primarily due to misaligned ground truth and imprecise part isolation, as depicted in Figure \ref{fig:failures}. Ground truth often mismatched the actual object parts, and VLM occasionally segmented entire objects instead of a part. The system exhibited weaknesses at multiple stages:

\begin{figure}
    \centering
    \includegraphics[width=0.95\linewidth]{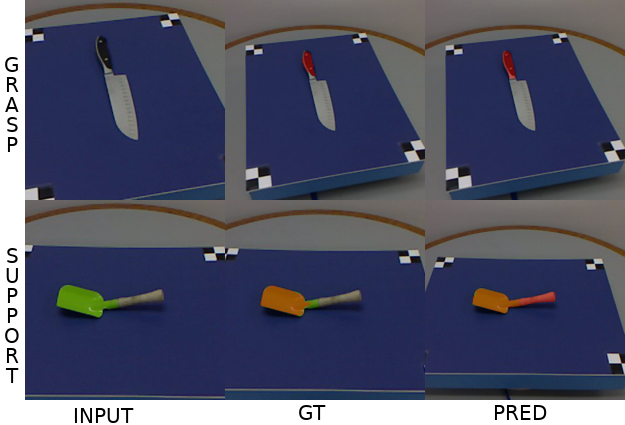}
    \caption{Segmentation Failures}
    \label{fig:failures}
\end{figure}
\begin{itemize}
    \item Class names: VLM struggled with class names like ``turner" and ``pot," causing confusion. Using more common names such as ``spatula" instead of ``turner" and ``plant pot" for ``pot" might improve its performance, as VLM better recognizes common names.
    \item Spatial relations: The LLM produced part descriptions involving spatial relationships (e.g. cup top), which were difficult for the VLM to process. To overcome this, the LLM was prompted once more for alternative names when the VLM failed to recognize the initially described part. This approach improved the VLM's detection accuracy when the reprompted part was a physical part (e.g. cup rim). %
\end{itemize}

\subsection{Robot Demonstration}

\begin{figure}[hbt!]
    \centering
    \includegraphics[width=0.8\linewidth]{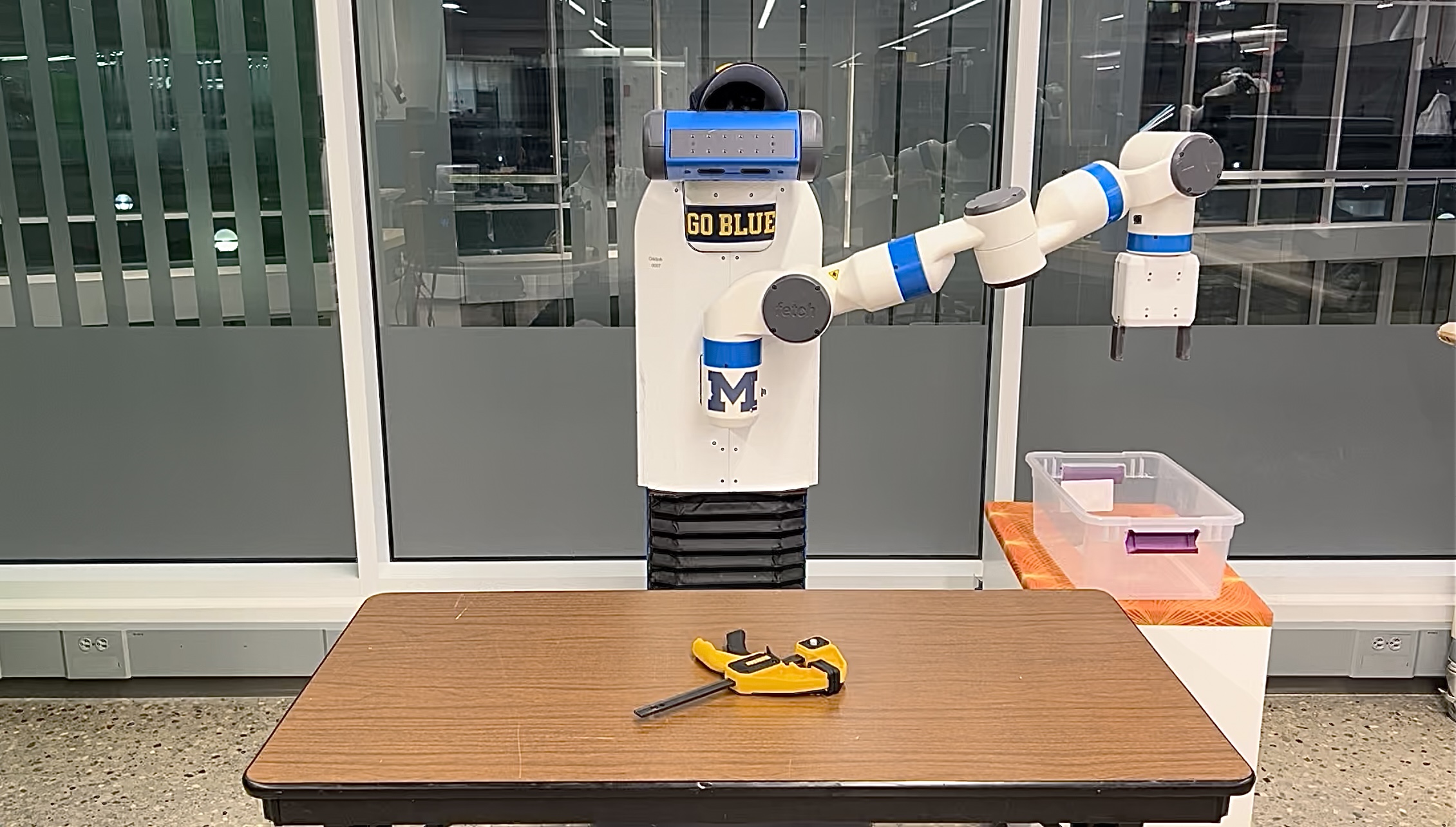}
    \caption{Robot Experimental Setup \\ Video: https://youtu.be/fXvYo-0AJek}
    \label{fig:robot experiment}
\end{figure}

To demonstrate the concept with an actual robot, we utilized a Fetch and Freight Research Edition robot to grasp various random objects that were placed on a table directly in front of it. We captured both RGB and depth images using the onboard cameras. The RGB image was then processed through the previously described pipeline, where we segmented the image area that affords grasping, resulting in a binary mask. This mask, along with the depth information, was fed into DexNet\cite{mahler2017dexnet} to suggest a potential grasp. To simplify the grasping process, we modified the grasp suggested by DexNet\cite{mahler2017dexnet} and converted it into a top-down pick-up. This was achieved by using the x, y, z coordinates, setting the gripper orientation to face downwards along the negative z-axis, and then calculating the z-axis rotation for the wrist rotation. The finalized grasp configuration was then executed by the robot using the MoveIt\cite{moveit}. A image of the setup of shown in figure \ref{fig:robot experiment}.

The experiment involved a spatula, walkie-talkie, toy lightsaber, mug, clamp, and brush, repeated ten times each. Table \ref{tab:grasping_metrics} shows successful segmentations where segments matched the correct object parts, the rate of feasible grasp proposals at suggested parts, and successful object pickups by the robot at these parts.

\begin{table}[h!]
  \centering
    \caption{Grasping Performance Metrics}

  \begin{tabular}{llccc}
    \hline
    Object &  Sugge- & Segmentation  & Grasp  & Pick  \\
    & sted Part & Success & Success & Success \\
    \hline
    Brush & Handle & 10/10 & 10/10 & 10/10 \\
    Clamp & Handle & 6/10 & 6/6 & 5/6 \\
    Mug & Handle & 10/10 & 10/10 & 7/10 \\
    Lightsaber & Handle & 5/10 & 5/5 & 5/5 \\
    Spatula & Handle & 7/10 & 7/7 & 6/7 \\
    Walkie Talkie & Body & 4/10 & 4/4 & 4/4 \\
    \hline
  \end{tabular}
  \label{tab:grasping_metrics}
\end{table}

Failures are mostly caused by imprecise segmentation that include whole objects instead of specific parts. However, with accurate segmentation, grasp and pickup success rates approach 100\%. Grasping failures occurred for the mug because its shapes was not considered in grasp planning.

\subsection{Ablation}
To validate our method, we removed components from our network to evaluate their impact on performance. We tested two variations: 1) using only the VLM with a list of affordances and 2) excluding the reprompting step.

\begin{table}[h!]
\centering
\caption{Ablation on UMD dataset}

\begin{tabular}{lccc}
\hline
 & VLM only & No Reprompting & Full Network \\ \hline
grasp & 0.002 & 0.498 & 0.650 \\ 
cut & 0.013 & 0.401 & 0.823 \\ 
scoop & 0.062 & 0.345 & 0.753 \\
contain & 0 & 0.437 & 0.688 \\ 
pound & 0 & 0.317 & 0.809 \\ 
support & 0 & 0.436 & 0.538 \\ 
wrap-grasp & 0 & 0.312 & 0.718 \\ \hline
Average & 0.011 & 0.392 & 0.711 \\ \hline
\end{tabular}
\label{table:ablation}
\end{table}

Table \ref{table:ablation} shows that a standalone VLM struggles with affordances, likely because VLMs are not typically trained on such tasks. The necessity of reprompting for alternative part names is also highlighted, emphasizing how the VLM's open vocabulary can respond variably to different words, showing that using synonyms can improve results.

%% file: secs/6_conclusion.tex
\section{Conclusion and Future Work}
Our experiments demonstrate OVAL-Prompt's potential for open-vocabulary affordance localization, achieving results in the UMD dataset that is comparable to supervised models without fine-tuning. Furthermore, OVAL-Prompt's efficacy in real-world robotic tasks highlight its practicality. 

Our method efficiently performs zero-shot affordance localization but struggles with scalability for numerous items due to individual LLM queries. Future improvements could enhance simultaneous multi-item and affordance identification. Testing in cluttered environments and eliminating the VLM's need for predefined object lists are also potential development areas.

%% file: secs/7_supplementary.tex
\section{Supplementary Material}
In addition to the UMD dataset we evaluated our method on the AGD20K dataset.
\subsubsection{AGD20K}
The AGD20K dataset is intended for affordance grounding tasks and includes over 20,000 images across 36 affordance categories. It utilizes three evaluation metrics: Kullback-Leibler Divergence (KLD), similarity metric (SIM), and Normalized Scanpath Saliency (NSS). Detailed information on their design and implementation can be found in the supplementary material of the Luo et al. paper \cite{Learningluo}. For the performance of other models, we referenced the data from the paper by Li et al. \cite{li2023locate}.

Our method (see table \ref{tab:AGD20K}) has higher KLD than the best seen model by 9.429 points and the unseen by 7.427. Our SIM and NSS scores were competitive with other models beating several models and only being surpassed by LOCATE\cite{li2023locate} for both seen and unseen categories, suggesting our segmentation closely aligns with the ground truth. The high KLD stems from our method producing a binary mask for the affordance area, in contrast to the ground truth comparison that uses a value distribution, the peak of which indicates the affordance region. Consequently, our non-distributional output incurs a significant penalty in KLD calculation.

\begin{table}[!h]
\centering
\caption{AGD20K Dataset Evaulation}
\begin{tabular}{lcccccc}
\hline
{\textbf{Method}} & \multicolumn{3}{c}{\textbf{Seen}} & \multicolumn{3}{c}{\textbf{Unseen}} \\
& \textbf{KLD↓} & \textbf{SIM↑} & \textbf{NSS↑} & \textbf{KLD↓} & \textbf{SIM↑} & \textbf{NSS↑} \\
\hline
\textbf{EIL\cite{Mai_2020_CVPR} } & 1.931 & 0.285 & 0.522 & 2.167 & 0.227 & 0.330 \\
\textbf{SPA\cite{Pan_2021_CVPR} } & 5.528 & 0.221 & 0.357 & 7.425 & 0.169 & 0.262 \\
\textbf{TS-CAM \cite{ts_cam}} & 1.842 & 0.260 & 0.336 & 2.104 & 0.201 & 0.151 \\
\textbf{Hotspots \cite{nagarajan2019grounded} } & 1.773 & 0.278 & 0.615 & 1.994 & 0.237 & 0.577 \\
\textbf{Cross-view-AG \cite{Learningluo}} & 1.538 & 0.334 & 0.927 & 1.787 & 0.285 & 0.829 \\
\textbf{Cross-view-AG+ \cite{luo2023grounded}} & 1.489 & 0.342 & 0.981 & 1.765 & 0.279 & 0.882 \\
\textbf{AffCorrs†\cite{hadjivelichkov2022oneshot}} & 1.407 & 0.359 & 1.026 & 1.618 & 0.348 & 1.021 \\
\textbf{LOCATE\cite{li2023locate}} & \textbf{1.226} & \textbf{0.401} & \textbf{1.177} & \textbf{1.405} & \textbf{0.372} & \textbf{1.157} \\
\textbf{OVAL-Prompt(Ours)} & 10.649 & 0.339 & 1.044 & 8.832 & 0.365 & 0.925\\
\hline
\end{tabular}
\label{tab:AGD20K}
\end{table}